\documentclass[twoside,reqno]{amsart}
\usepackage{oooaistats2016}

\usepackage{graphicx}
\usepackage{url}
\usepackage{booktabs}
\usepackage{amsmath}
 \usepackage{amsthm}
\usepackage{amssymb}
\usepackage{bbm}
\usepackage{enumerate}
\usepackage{multirow}
\usepackage{soul,color}
\usepackage{times}

\usepackage{array}
\usepackage{bm}
\usepackage{color}
\usepackage{enumerate}
\usepackage{natbib}

\usepackage{amsthm}
 
\theoremstyle{definition}

\theoremstyle{theorem}
\newtheorem{theorem}{Theorem}
\theoremstyle{example}

\theoremstyle{proposition}
\newtheorem{proposition}{Proposition}

\begin{document}

% If your paper is accepted and the title of your paper is very long,
% the style will print as headings an error message. Use the following
% command to supply a shorter title of your paper so that it can be
% used as headings.
%
%\runningtitle{I use this title instead because the last one was very long}

% If your paper is accepted and the number of authors is large, the
% style will print as headings an error message. Use the following
% command to supply a shorter version of the authors names so that
% they can be used as headings (for example, use only the surnames)
%
%\runningauthor{Surname 1, Surname 2, Surname 3, ...., Surname n}

\twocolumn[

\aistatstitle{State Space representation of non-stationary Gaussian Processes}

\aistatsauthor{ Alessio Benavoli \And Marco Zaffalon\\ }

  \aistatsaddress{alessio@idsia.ch \hspace{4.7cm}  zaffalon@idsia.ch \\ ~\\
  Dalle molle institute for artificial intelligence (IDSIA),\\ Manno, Switzerland} ]

\begin{abstract}
The state space (SS) representation of Gaussian processes (GP) has recently gained a lot of interest.
The main reason is that it allows to compute GPs based 
inferences in $\mathcal{O}(n)$, where $n$ is the number of observations.
This implementation  makes GPs suitable for Big Data.  For this reason, it is important to
provide a SS representation of the most important kernels used in machine learning.
The aim of this paper is to show how to exploit the transient behaviour of SS models to 
map non-stationary kernels to SS models.
\end{abstract}

% \section{GENERAL FORMATTING INSTRUCTIONS}

% IDEE:
% \begin{itemize}
%  \item inserire integrale nella risposta per un esempio
% \end{itemize}

\section{Introduction}
In machine learning, \textit{Gaussian Processes} (GP) are commonly used modelling tools
for Bayesian non-parametric inference \citep{Hagan1978,bernardo1998regression,mackay1998introduction,rasmussen2006gaussian,GPsite,gelman2013bayesian}.
For instance in GP regression, $y=f(x)$, the aim is to estimate $f$ from (noisy) observation $y$. A natural Bayesian way to approach this problem is to place a prior on $f$ and use the observations to compute  the posterior of $f$.
Since $f$ is a function, the GP is a natural prior distribution for $f$ \citep{mackay1998introduction,rasmussen2006gaussian}. 
A GP, denoted as $GP(0,k_{\bm{\theta}}(x,x'))$, is completely defined by its mean function (usually assumed to be zero)
and covariance function (CF) (also called kernel) $k_{\bm{\theta}}(x,x')$, which depends
on a vector of hyperparameters $\bm{\theta}$. 
By suitably choosing the kernel function, we can make GPs very flexible and convenient modelling tools.
However, a drawback with GPs is that the direct computation of the posterior of $f$ is computationally demanding. 
The computational cost is cubic, $\mathcal{O}(n^3)$, in the number of observations.  
This makes GPs unsuitable for Big Data. Several general sparse approximation schemes
have been proposed for this problem, (see for instance \cite{quinonero2005unifying} and \cite[Ch. 8]{rasmussen2006gaussian}).

In the case the function $f$ is defined on $\mathbb{R}$,  i.e., $x\in \mathbb{R}$, computational savings
can be made by converting the GP into \textit{State Space} (SS)
form and make inference using \textit{Kalman filtering}. 
Note that the case $x\in \mathbb{R}$ is particularly important because it includes time-series analysis ($x$ is time).
The connection between GPs and SS models is well known for some basic 
kernels  and recently it has gained a lot of interest. Certain
classes of stationary CFs can be directly converted into state space models by representing their spectral densities 
as rational functions \citep{sarkka2012infinite,sarkka2013spatiotemporal,solin2014gaussian}.
 Moreover, an explicit link between periodic (non-stationary)  CFs and SS models has also been derived in \citep{solin2014explicit}.

The connection between the SS representation of GPs and GPs models used in  machine learning  is
important for many reasons. 
First, in machine learning, it has been shown that GPs, with a suitable choice of the kernel, are universal function approximators
\citep{rasmussen2006gaussian,williams1997computing}. Moreover, GPs can also be used for classification, with only a slight modification.
Second, inferences in SS models can be computed efficiently  by processing
the observations sequentially. This means that the computational cost of inference in the SS representation of GPs
is  $\mathcal{O}(n)$.
Third, SS models represent GPs through  Stochastic Differential Equations (SDE). They return a model that directly explains the time-series and not only
fits or predicts it. It is well known that SDEs are basic modelling tools in econometrics, physics etc..
% Instead, the machine learning representation of GPs cannot directly be used to explain.
Hence, if we are able to map the most important kernels used in machine learning to the SS representation we can  ``kill three birds with one stone'', i.e.,
we can have an explanatory model with a universal function approximation property  at a cost of $\mathcal{O}(n)$.
This is the aim of this paper. In particular,  the goal is to extend the work \citep{sarkka2012infinite,sarkka2013spatiotemporal} by providing a SS representation of the most important kernels used in machine learning.

In particular, we will show that  non-stationary kernels can be mapped into SS models by considering the \textit{transient} behaviour. 
It is well known that the time response of linear SS models is always the superposition of an initial-condition part and a driven part.
The response due to initial-conditions is often ignored, because it vanishes in the stationary case (it is transient). However, for non-stationary  systems,
the transient never vanishes and, thus, it determines the behaviour of the system. Even with a zero initial condition, we can have a transient
behaviour due to the driven part. We will show that by taking into account the transient, we can map the linear regression, periodic and spline kernel to SS models. 
Moreover, we will also study the transient behaviour for stationary systems to show that in this case it vanishes.
To reconcile these two cases, we will make use of the Laplace transform that is able to account for the transient behaviour.
This is a difference w.r.t.\ the work by \cite{sarkka2012infinite,sarkka2013spatiotemporal} where they employed the Fourier transform.
Then we will show how to map the neural networks kernels to SS models. For this purpose we will use linear time-variant SS, that are intrinsically
non-stationary. Finally, by means of simulations we will show the effectiveness of the proposed approach and the computational advantages by applying it
to long time-series.
In this work, for lack of space, we will assume that the reader is familiar with the machine learning  representation of GPs and we will only discuss 
the SS representation.

\section{State Space model}
Let us consider the following stochastic linear time-variant (LTV) state space model \citep{jazwinski2007stochastic}
\begin{equation}
\label{eq:sys}
 \left\{\begin{array}{rcl}
d \mathbf{f}(t)&=&\mathbf{F}(t)\,\mathbf{f}(t)dt+\mathbf{L}(t)\,dw(t),\\
 y(t_k)&=&\mathbf{C}(t_k)\,\mathbf{f}(t_k), %+\epsilon_k,~~~\epsilon_k\sim\mathcal{N}(0,\sigma^2),
\end{array}\right.
\end{equation} 
where $\mathbf{f}(t)=[f_1(t),\dots,f_m(t)]^T$ is the (stochastic) state vector, $y(t_k)$
is the observation at time $t_k$, $w(t)$ is a one-dimensional Wiener process  with intensity $q(t)$ and $\mathbf{F}(t),\mathbf{L}(t),\mathbf{C}(t)$
are known time-variant matrices of appropriate dimensions.
We further assume that the initial state $\mathbf{f}(t_0)$ and $w(t)$ are independent for each $t\geq t_0$.
It is well know that the solution of the  stochastic differential equation in (\ref{eq:sys}) is (see for instance  \cite{jazwinski2007stochastic}):
\begin{equation}
\label{eq:sol}
\mathbf{f}(t_k)=\boldsymbol{\psi}(t_k,t_0)\,\mathbf{f}(t_0)+\int\limits_{t_0}^{t_k} \boldsymbol{\psi}(t_k,\tau)\mathbf{L}(\tau)\,dw(\tau),\\
\end{equation} 
with $\boldsymbol{\psi}(t_k,t_0)= \exp(\int_{t_0}^{t_k}  \mathbf{F}(t)dt)$ is the state transition matrix,
which is obtained as a matrix exponential.\footnote{The matrix exponential is $e^A=I+A+A^2/2!+A^3/3!+\dots$.}
\begin{proposition}
\label{prop:1}
 Assume that $E[\mathbf{f}(t_0)]=\mathbf{0}$, then  the vector 
of observations $[y(t_1),y(t_2),\dots,y(t_n)]^T$ is Gaussian distributed with zero mean
and covariance matrix whose elements are given by:
\begin{equation}
\label{eq:solcov}
\begin{array}{l}
E[y(t_i)y(t_j)]=\vspace{1mm}\\
% \mathbf{C}(t_i)\boldsymbol{\psi}(t_i,t_0)E[\mathbf{f}(t_0)\mathbf{f}^T(t_0)] (\mathbf{C}(t_j)\boldsymbol{\psi}(t_j,t_0))^T+\vspace{1mm}\\
% \int\limits_{t_0}^{t_i} \int\limits_{t_0}^{t_j}  \mathbf{C}(t_i) \boldsymbol{\psi}(t_i,u)\mathbf{L}(u)\,E[dw(u)dw(v)]\mathbf{L}^T(v)\boldsymbol{\psi}^T(t_j,v)\mathbf{C}^T(t_j) \vspace{1mm}\\
% =\mathbf{C}(t_i)\boldsymbol{\psi}(t_i,t_0)E[\mathbf{f}(t_0)\mathbf{f}^T(t_0)] (\mathbf{C}(t_j)\boldsymbol{\psi}(t_j,t_0))^T+\vspace{1mm}\\
% \int\limits_{t_0}^{\min(t_i,t_j)}  \mathbf{C}(t_i) \boldsymbol{\psi}(t_i,u)\mathbf{L}(u)\mathbf{L}^T(u)\boldsymbol{\psi}^T(t_j,u)\mathbf{C}^T(t_j)q(u)du\vspace{1mm}\\
=\mathbf{C}(t_i)\boldsymbol{\psi}(t_i,t_0)E[\mathbf{f}(t_0)\mathbf{f}^T(t_0)] (\mathbf{C}(t_j)\boldsymbol{\psi}(t_j,t_0))^T+\vspace{1mm}\\
\int\limits_{t_0}^{\min(t_i,t_j)} h(t_i,u)h(t_j,u)q(u)du
\end{array}
\end{equation} 
where we have exploited the fact that $E[dw(u)dw(v)]=q(u)\delta(u-v)dudv$ and defined $h(t_1,t_2)=\mathbf{C}(t_1) \boldsymbol{\psi}(t_1,t_2)\mathbf{L}(t_2)$ ($h(\dots)$ is called impulse response).
\end{proposition}
The proof of this proposition is well known (see for instance \cite{jazwinski2007stochastic}), but we have reported the derivations of this proposition (and next propositions/theorems) in appendix
for the convenience of the reader.
From  (\ref{eq:solcov}), it is evident that, given the time-varying matrices  $\mathbf{A}(t),\mathbf{L}(t),\mathbf{C}(t) $, the CF of a LTV system is completely defined by:
(i) the covariance of the initial condition $E[\mathbf{f}(t_0)\mathbf{f}^T(t_0)]$; (ii) the CF of the noise $E[dw(u)dw(u)]$.

% \subsection{Time invariant case and Laplace transform}
In case the SS model is \textit{Linear Time-Invariant} (LTI), i.e.,
$\mathbf{F}(t)=\mathbf{F},\mathbf{L}(t)=\mathbf{L},\mathbf{C}(t)=\mathbf{C},q(t)=q$, we
can use the \textit{Laplace transform} to derive (\ref{eq:sol}) and (\ref{eq:solcov}) using only algebraic computations. 
The Laplace transform of a function $x(t)$, defined for  $t\geq 0$, is:
$$
x(s) =\int\limits_{0}^\infty e^{-st} x(t)\, dt,
$$
where the parameter $s$ is the complex number $s = \sigma + \iota \omega$, with $\sigma,\omega \in \mathbb{R}$
and $\iota$ denoting the imaginary unit.
The Laplace transform exists provided that the above integral is finite. The values of $s$ for which the Laplace transform
exists are called the \textit{Region Of Convergence }(ROC) of the Laplace transform.
By using the Laplace transform, we can rewrite the differential equation in (\ref{eq:sys}) 
in an algebraic form:
\begin{equation}
\label{eq:syslaplace}
s \mathbf{f}(s)-\mathbf{f}(t_0) =\mathbf{F} \mathbf{f}(s)+\mathbf{L}\,w(s),
\end{equation} 
where $\mathbf{f}(s),w(s)$  are the Laplace transforms of $\mathbf{f}(t),w(t)$.\footnote{By defining the Laplace transform (\ref{eq:syslaplace}), we are wrongly considering $w(t)$
as a deterministic input. We use this notation only for convenience, but then we define the correct inverse Laplace transform in (\ref{eq:sysinvlaplace}).}
Since $y(s)=\mathbf{C}\mathbf{f}(s)$, we have
\begin{align}
\nonumber
 y(s)&=\mathbf{C}(s\mathbf{I}-\mathbf{F})^{-1}\mathbf{f}(t_0)+\mathbf{C}(s\mathbf{I}-\mathbf{F})^{-1}\mathbf{L}\,w(s)\\
&=\mathbf{C}(s\mathbf{I}-\mathbf{F})^{-1}\mathbf{f}(t_0)+H(s)\,w(s),
\end{align}
where $H(s)=\mathbf{C}(s\mathbf{I}-\mathbf{F})^{-1}\mathbf{L}$ is called the transfer function of the linear time-invariant (LTI) SS model
and $\mathbf{I}$ is the identity matrix.
Since a product in the Laplace domain corresponds to a convolution in time, it follows that
\begin{equation}
\label{eq:sysinvlaplace}
y(t_k)=\mathcal{L}^{-1}\left(\mathbf{C}(s\mathbf{I}-\mathbf{F})^{-1}\right)\mathbf{f}(t_0)+ \int_{t_0}^{t_k} h(t_k-u)dw(u), 
\end{equation} 
where $\mathcal{L}^{-1}(\cdot)$ denotes the inverse Laplace transform. By computing $E[y(t_i)y(t_j)]$, we obtain again
(\ref{eq:solcov}).
The output of both LTV and LTI systems is clearly completely defined by the SS matrices, the initial condition
and the stochastic forcing term $dw(t)$.
The aim of the next sections is to show that by suitably choosing these three components we can 
obtain SS models whose CF coincides with the main kernels used in GPs.

\section{Non-stationary CFs defined by LTI SS without forcing term}
In this section, we will show that two important CFs used in GPs  can be obtained by two LTI SS models without stochastic forcing term. Their output is therefore completely determined by the initial conditions and, thus, the CF
they define is non-stationary (it depends on  $t_i,t_j\geq t_0$).
 
\subsection{Linear regression kernel}
Assume without loss of generality that $t_0=0$, 
\begin{align}
\nonumber
\mathbf{F}=& \left[\begin{array}{cc}
0 & 1\\
0 & 0\\
 \end{array}\right],~\mathbf{L}= \left[\begin{array}{c}
0 \\
1 \\
 \end{array}\right],\\
\label{eq:sysreg}
 \mathbf{C}=& \left[\begin{array}{cc}
1 & 0 \\
 \end{array}\right], 
 \end{align} 
 $q(t)=0 ~\forall~ t\geq 0$\footnote{Since $q(t)=0$, the matrix $\mathbf{L}$ is completely superfluous. We have introduced it only because we will use this model later.} and so there is not forcing term ($dw(t)=0$).  This corresponds to the following LTI SS model:
\begin{equation}
\label{eq:syslr}
 \left\{\begin{array}{rcl}
\frac{d f_1}{dt}(t)&=&f_2(t),\vspace{1mm}\\
\frac{d f_2}{dt}(t)&=&0, \vspace{1mm}\\
 y(t_k)&=&f_1(t_k). %+\epsilon_k,~~~\epsilon_k\sim\mathcal{N}(0,\sigma^2),
\end{array}\right.
\end{equation}
From the equations of this SS model, since $y(t)=f_1(t)$ we derive that $f_1(t)=f(t)$ (it is the function 
of interest) and  $f_2(t)=\frac{d f}{dt}(t)$ is its derivative. 
% This is particularly interesting: SS  models the function and the derivatives of the function.
% This may be used to perform monotonicity tests on $f$ in a similar way to  \citep{benavoli2015a}
% using the ordinary representation of GPs.
By computing $\boldsymbol{\psi}(t,t_0)= \exp(\int_{t_0}^{t}  \mathbf{F}dt)=\exp((t-t_0) \mathbf{F})$, we have that
$$
\boldsymbol{\psi}(t,t_0)=\left[\begin{array}{cc}
1 & t-t_0\\
0 & 1\\
 \end{array}\right],
$$
and, therefore, $y(t_i)=\mathbf{C}\boldsymbol{\psi}(t_i,t_0)\mathbf{f}(t_0)=f_1(t_0)+(t-t_0) f_2(t_0)$.
\begin{proposition}
\label{prop:2}
Consider the SS model in (\ref{eq:syslr}) and assume that $\mathbf{f}(t_0)$ is Gaussian distributed with zero mean and covariance $E[\mathbf{f}(t_0)\mathbf{f}^T(t_0)]=diag([\sigma_1^2,\sigma_2^2])$ and $t_0=0$. From (\ref{eq:sol}),
we have that $[y(t_1),\dots,y(t_k)]$ is Gaussian distributed with zero mean and CF 
$$
E[y(t_i)y(t_j)]=\sigma_1^2+\sigma_2^2t_it_j,
$$
 for each $t_i,t_j\geq 0$, which corresponds to the linear regression CF \cite[Sec. 4.2.2]{rasmussen2006gaussian}. 
\end{proposition}
This connection between SS and linear regression is well-known, here we have shown how  to derive the  CF.
Higher order linear regression CFs can  be obtained
by considering   $E[\mathbf{f}(t_0)\mathbf{f}^T(t_0)]=diag([\sigma_1^2,\dots,\sigma_m^2])$
and 
\begin{align}
\nonumber
\mathbf{F}=& \left[\begin{array}{ll}
\mathbf{0}_{m-1} & \mathbf{I}_{m-1}\\
0 & \mathbf{0}_{m-1}^T\\
 \end{array}\right],~\mathbf{L}= \left[\begin{array}{l}
\mathbf{0}_{m-1}^T \\
1 \\
 \end{array}\right]\\
\label{eq:sysregho}
 \mathbf{C}=& \left[\begin{array}{cc}
\mathbf{0}_{m-1} & 1 \\
 \end{array}\right], 
 \end{align} 
 where $\mathbf{0}_{m-1}$ is the $(m-1)$ zero vector.

\subsection{Periodic Kernel}
Consider the following LTI SS:
  \begin{align}
\label{eq:syspe}
\mathbf{F}=& \left[\begin{array}{cc}
0  & \omega_k\\
-\omega_k & 0\\
 \end{array}\right], \mathbf{C}= \left[\begin{array}{cc}
1  & 0 \\
 \end{array}\right], 
 \end{align} 
 with $q(t)=0$ for all $t\geq 0$.
 By computing $\boldsymbol{\psi}(t,t_0)= \exp((t-t_0) \mathbf{F})$, we have that
$$
\boldsymbol{\psi}(t,t_0)=\left[\begin{array}{cc}
~~\cos(\omega_k (t-t_0)) & \sin(\omega_k (t-t_0))\\
-\sin(\omega_k (t-t_0)) & \cos(\omega_k (t-t_0))\\
 \end{array}\right],
$$
and, so
  \begin{align}
  \nonumber
  y(t_i)&=\mathbf{C}\boldsymbol{\psi}(t_i,t_0)\mathbf{f}(t_0)\\
\nonumber
&=\cos(\omega_k (t-t_0))f_1(t_0)+ \sin(\omega_k (t-t_0))f_2(t_0)\\
% &=(\cos(\omega_k (t-t_0)) - \sin(\omega_k (t-t_0)))f_1(t_0)\\
% \nonumber
% &+(\cos(\omega_k (t-t_0)) + \sin(\omega_k (t-t_0)))f_2(t_0)\\
\label{eq:syspeout}
&=a_k\cos(\omega_k (t-t_0))+b_k \sin(\omega_k (t-t_0)),
 \end{align}
 where we have written $f_1(t_0)=a_k$ and  $f_2(t_0)=b_k$.
 \begin{proposition}
\label{prop:3}
Consider the SS model in (\ref{eq:syspe}) and assume that $[a_k,b_k]$ are Gaussian distributed with zero mean, 
variances $E[a_k^2]$, $E[b_k^2]$ and uncorrelated  $E[a_kb_k]=0$. From (\ref{eq:sol}),
we have that $[y(t_1),\dots,y(t_n)]$ is Gaussian distributed with zero mean and CF 
  \begin{align}
\nonumber
E[y(t_1)y(t_2)]&=E[a_k^2]\cos(\omega_k (t_1-t_0))\cos(\omega_k (t_2-t_0))\\
\nonumber
&+E[b_k^2]\sin(\omega_k (t_1-t_0))\sin(\omega_k (t_2-t_0)).\\
 \nonumber
 &=E[a_k^2]\cos(\omega_k (t_2-t_1))
 \end{align} 
 for each $t_i,t_j\geq t_0$, where last equality holds if $E[a_k^2]=E[b_k^2]$.
\end{proposition}
This SS model defines a periodic CF. However,
it is evident (see in particular (\ref{eq:syspeout})) that it can only represent sinusoidal  type periodic functions.
However, we know that any periodic function $f(t)$ in $[-p_e,p_e]$,  that is integrable, can be approximated by 
 Fourier series:
\begin{align} 
\label{eq:fourierser}
f(t) &\approx \sum_{k=1}^J \left[a_k\cos\left(\frac{2 \pi k t}{p_e}\right)+b_k\sin\left(\frac{2 \pi k t}{p_e}\right)\right], 
\end{align}
where $a_k,b_k\in \mathbb{R}$ are the coefficients of the Fourier series, which depend on $f(t)$,
and $J\in \mathbb{N}$ is the order of the approximation.\footnote{We can also include the constant term ($k=0$)
in the  series.}
Therefore, we can approximate any periodic function by a sum of $J$ SS models 
 of type (\ref{eq:syspe}) with $\omega_k=\tfrac{2 \pi k}{p_e}$.
 For instance, for $J=2$, we can consider the SS model
   \begin{align}
\label{eq:syspext}
\mathbf{F}=& \left[\begin{array}{cccc}
0  & \omega_1 & 0 &0\\
-\omega_1 & 0 & 0 & 0\\
0 & 0 & 0 & \omega_2  \\
0 & 0 & -\omega_2 & 0 \\
 \end{array}\right], \mathbf{C}= \left[\begin{array}{cccc}
1  & 0 & 1 &0\\
 \end{array}\right]. 
 \end{align} 
 Its  transfer function $\phi(t_i,t_0)$ is a diagonal block matrix with blocks
$$
\left[\begin{array}{cc}
~~\cos(\omega_k (t-t_0)) & \sin(\omega_k (t-t_0))\\
-\sin(\omega_k (t-t_0)) & \cos(\omega_k (t-t_0))\\
 \end{array}\right],
$$
 for $k=1,2$.
 Hence, from (\ref{eq:solcov}), we have that
  \begin{equation}
\label{eq:solcovi01}
\begin{array}{l}
E[y(t_i)y(t_j)]=\sum\limits_{k=1}^2 E[a_k^2]\cos(\omega_k (t_i-t_0))\cos(\omega_k (t_j-t_0))\\
+E[b_k^2]\sin(\omega_k (t_i-t_0))\sin(\omega_k (t_j-t_0)),\\
\end{array}
\end{equation}
where we have assumed that $E[\mathbf{f}(t_0)\mathbf{f}^T(t_0)]$ is a block diagonal matrix with blocks  
  \begin{equation}
\label{eq:blovkfour}
\left[\begin{array}{cc}
E[a_k^2] & 0\\
0& E[b_k^2]
 \end{array}\right],
\end{equation}
for $k=1,2$. 
In Fourier series, the function $f(t)$ is known and so the coefficients $a_k,b_k$ can be computed based on $f(t)$.
In GP regression, we do not know $f(t)$ and so we do not know $a_k,b_k$. We must estimate these
coefficients from data.\footnote{When the period $p_e$ is unknown, we can estimate it from data.}
This is the reason we have assumed a prior distribution on these coefficients. 
This prior is completely defined by the variances $E[a_k^2],E[b_k^2]$ for $k=1,2,\dots,J$.
If we further assume that $E[a_k^2]=E[b_k^2]=q_k^2$ then we have only $J$ parameters to specify, the $q_k$.
We can further assume that  all the parameters are functions of a single parameter $\ell \geq 0$;
and penalize high order frequencies. For instance, if we choose
$q_k^2=2\exp(-\ell^{-2}) \sum_{i=0}^{\lfloor \frac{J- k}{2}\rfloor} \frac{(2 \ell^2)^{-k-i}}{(k+i)!i!}$,
it can be shown that
  \begin{align}
\nonumber
&E[y(t_i)y(t_j)]=\sum_{k=0}^J q_k^2\cos\left(\frac{2 \pi k}{p_e} (t_j-t_i)\right)\\
\nonumber
&\xrightarrow{J \rightarrow \infty} \exp\left(-\tfrac{2}{\ell^2}\sin\left(\frac{\pi(t_j-t_i)}{p_e}\right)^2\right),
 \end{align} 
 which is the periodic CF used in GPs \citep[Sec. 4.2.3]{rasmussen2006gaussian}.
 This result has been derived by \cite{solin2014explicit}, here we have highlighted more extensively the transient analysis and the connection with the Fourier series.
%  In other words, (\ref{eq:fourierser})

\section{Non-stationary CFs defined by LTI SS with zero initial conditions}
In this section, we will show that an important CF used in GPs  can obtained by a LTI SS model with 
stochastic forcing term and zero initial conditions.

\subsection{Spline kernel}
We derive the CF for the cubic smoothing splines.
 Consider the LTI SS model in (\ref{eq:sysreg}), but this time
assume that $q(t)=1 ~\forall~ t\geq t_0$,
$E[\mathbf{f}(t_0)\mathbf{f}^T(r_0)]=\mathbf{0}$ and $t_0=0$, then
\begin{align}
\nonumber
 y(t_i)&= \int\limits_{0}^{t_i} \mathbf{C} \boldsymbol{\psi}(t_i,\tau)\mathbf{L}(\tau)\,dw(\tau)= \int\limits_{0}^{t_i} (t_i-\tau)\,dw(\tau)\\
\end{align}
and so, from (\ref{eq:sol}),
$ E[y(t_i)y(t_j)] = \int_{0}^{\min(t_i,t_j)} (t_i-\tau)(t_j-\tau)\,d\tau$.
\begin{proposition}
\label{prop:4}
 Consider the LTI SS model in (\ref{eq:sysreg}), but this time
assume that $q(t)=1 ~\forall~ t\geq 0$ and
$E[\mathbf{f}(0)\mathbf{f}^T(0)]=\mathbf{0}$.
Then, from (\ref{eq:sol}), we derive that $[y(t_1),\dots,y(t_n)]$ is Gaussian distributed with zero mean and CF  
\begin{align}
\label{eq:covsplines}
 E[y(t_i)y(t_j)]&=|t_i-t_j|\frac{\min(t_i,t_j)^2}{2}+\frac{\min(t_i,t_j)^3}{3},
\end{align}
for each $t_i,t_j\geq 0$, which is the kernel for the cubic smoothing splines \citep[Sec. 6.3.1]{rasmussen2006gaussian}.
\end{proposition}
Higher order splines can be obtained considering SS models  as in (\ref{eq:sysregho}).
The connection between SS and smoothing splines has been first derived in  \cite{kohn1987new}.
Here, we have highlighted the connection with GPs and computed the CF (\ref{eq:covsplines}).

%  
% 
%  Then, it can be verified that
%  $$
%  h(z,u)= \mathbf{C}(z) \phi(z,u)\mathbf{L}=\frac{1}{\sqrt{\ell}}\exp\left(-\frac{z^2+u^2}{\ell^2}\right),
%  $$
%  and so
%  $$
% \begin{array}{l}
%  cov(y(z_1),y(z_2))\\
%  =\int_{z_0}^{\min(z_1,z_2)}\exp\left(-\frac{z_1^2+u^2}{\ell^2}\right)\exp\left(-\frac{z_2^2+u^2}{\ell^2}\right) du\\
%  \approx \frac{\sqrt{\frac{\pi }{2}} \left(
%    \text{erf}\left(\frac{\sqrt{2} \min(z_1,z_2)}{\ell}\right)+1\right)
%    e^{-\frac{z_1^2+z_2^2}{\ell^2}}}{2 }
%  \end{array}
%  $$
% Now define
% \begin{align}
% \nonumber
%  z_1&=(\cos(t'_2)-\cos(t'_1))^2\\
% \nonumber
%  z_2&=(\sin(t'_2)-\sin(t'_1))^2
% \end{align}
% with $t_i'=\pi t_i/p_e$, since
% $$
% \begin{array}{l}
% (\cos(t'_2)-\cos(t'_1))^2+(\sin(t'_2)-\sin(t'_1))^2\\
% =4\sin\left(\frac{\pi(t_1-t_2)}{p_e}\right)^2,
% \end{array}
% $$
% then 
% $$
%  e^{-\frac{z_1^2+z_2^2}{\ell^2}}=\exp\left(-\frac{4 \sin\left(\frac{\pi(t_1-t_2)}{p_e}\right)^2}{\ell^2}\right),
% $$
% which is  the periodic kernel (a nonstationary  kernel).

\section{Stationary CFs defined by LTI SS models}
A stationary CF is a CF that only depends on $t_j-t_i$.
In this section we will present  SS models whose CF corresponds to stationary kernels used in GPs.
Since stationary CFs satisfy 
$k(\tau)=k(-\tau)$ for  $\tau=t_j-t_i$, i.e., they are even functions defined on $\mathbb{R}$, it is convenient to introduce  
the bilateral Laplace transform:
$$
\mathcal{B}(f(\tau)) =\int\limits_{-\infty}^\infty e^{-s\tau} f(\tau)\, d\tau, 
$$
that is defined for functions that take values in $\mathbb{R}$. 
% The ROC is defined
% in a similar way to the (unilateral) Laplace transform.
When the ROC includes the imaginary axis, then for $s= \iota \omega$, the bilateral Laplace transform 
reduces to the Fourier transform.
Assume that the CF is stationary. The Wiener-Khintchine theorem \citep{chatfield2013analysis} states
that the CF is completely defined by its Fourier transform (its bilateral Laplace transform computed for $s=\iota \omega$)  (when it exists) and vice versa:
$$
k(\tau)=\int_{\mathbb{R}}S_y(\iota \omega)e^{2\pi \iota \omega \tau}d\omega,~~S_y(\iota \omega)=\int_{\mathbb{R}} k(\tau)e^{-2\pi \iota \omega \tau}d\tau,
$$
where $S_y(\iota  \omega)$ is the Fourier transform of $k(\tau)$ also called the \textit{spectral density}.
\begin{theorem}[Representation theorem]
\label{th:1}
Define $S_y(\iota  \omega)=S_y(s)$ and assume that $S_y(s)$ that is a (proper) rational function of $s$.
Then there exists a stable LTI system with the impulse response $h$ such that
\begin{align}
\label{eq:solcovlap00}
y(t)=\int_{-\infty}^t h(t-u)dw(u)
\end{align}
where $w$ is a stationary process with uncorrelated increments and spectral density $S_w(s)=q>0$.
In the Laplace domain this can be written as
\begin{align}
\label{eq:solcovlap}
S_y(s)=H(s)H(-s)q,
\end{align}
where $H(s)=b(s)/a(s)$ is a rational function whose denominator $a(s)$ has all roots with negative real parts, $b(s)$ has no roots with positive real part and  $q$ is a positive real constant.\footnote{Since $S_y(s)$ is a real even functions, the zeroes and poles are symmetric w.r.t.\ the real axis and mirrored in the imaginary axis.}
\end{theorem}
This is standard result for LTI systems.
By interpreting $q$ as the Bilateral Laplace transform of the noise $dw(t)$ ($S_w(s)=q$),
then  Equation (\ref{eq:solcovlap}) relates the spectral functions of the output of a  SS model
described by the transfer function $H(s)$ with that of the input (the noise $w(t)$) through the transfer function of the SS model $H(s)$.
We can use (\ref{eq:solcovlap}) to derive the SS model that is related to $S_y(s)$.
These are the steps: (i) find the zeroes and poles of $S_y(s)$; (ii) take all the poles $p_i$ with real negative part and zeroes with non-positive real part; (iii) decompose $H(s)$ as
 $$
 H(s)=\frac{\prod_i (s-z_i)}{\prod_i (s-p_i)}=\tfrac{b_0 s^n+b_{1}s^{n-1}+\dots+b_{n-1}s+b_n}{s^n+a_{1}s^{n-1}+\dots+a_{n-1}s+a_n},
 $$ 
 and derive the (observable canonical) LTI SS model:
 \begin{align}
\nonumber
\mathbf{F} &= \begin{bmatrix} 0& 1& 0& \cdots &0\\0 & 0& 1& \cdots & 0\\ \vdots & \vdots & 0&  \ddots & 0\\ 0& 0& \cdots & \vdots & 1\\-a_{n}& -a_{n-1}&  -a_{n-2}& \cdots & -a_1\\ \end{bmatrix}, \mathbf{L}=\begin{bmatrix} r_1\\ r_2\\ \vdots\\  r_{n-1}\\ r_{n} \end{bmatrix}\\
\label{eq:canon}
\mathbf{C} &= \begin{bmatrix} 1& 0& 0& \cdots & 0 \end{bmatrix}, \, 
 \end{align}
 where 
 $$
 \begin{bmatrix} r_0\\ r_1\\ \vdots\\   r_{n} \end{bmatrix}=\begin{bmatrix} 1 & & & &\\a_1 & 1 & & &\\ \cdots & & &\ddots &\\ a_n & \cdots & & a_1 &1\\ \end{bmatrix}^{-1}\begin{bmatrix} b_0\\ b_1\\ \vdots\\   b_{n} \end{bmatrix}
 $$
The above LTI system has CF   $k(\tau)$.

\subsection{Mat{\'e}rn kernel for $\nu=3/2$}
Let us consider again the stationary Mat{\'e}rn kernel for $\nu=3/2$, i.e., $E[y(t_i)y(t_j)]=k(\tau)=\tfrac{1}{4 \lambda^3} e^{-\lambda|\tau|}(1+ \lambda |\tau|)$ for $\tau=t_2-t_1$
\citep[Sec. 4.2.1]{rasmussen2006gaussian}. The bilateral Laplace transform of  $k(\tau)$  is
$$
\mathcal{B}\left(R(\tau)\right)=\frac{1}{(\lambda-s)^2(\lambda+s)^2},
$$
which is a rational function. We can then apply Theorem \ref{th:1}
and, in this case, $H(s)=1/(\lambda+s)^2$ and $S_w(s)=1$. 
Using (\ref{eq:canon}) we can derive the SS model:
\begin{align}
\nonumber
\mathbf{F}=& \left[\begin{array}{cc}
0  & 1\\
-\lambda^2 & -2\lambda\\
 \end{array}\right],~\mathbf{L}= \left[\begin{array}{c}
0 \\
1 \\
 \end{array}\right]\\
\label{eq:sysmatern}
 \mathbf{C}=& \left[\begin{array}{cc}
1 & 0 \\
 \end{array}\right], 
 \end{align} 
 $q(t)=1 ~\forall~ t\geq t_0$.
 Its impulse response is  $h(t)=t\exp(-\lambda t)$ for $t\geq t_0$.
 \begin{proposition}
 \label{prop:5}
  Consider (\ref{eq:sysmatern}) and assume that $E[\mathbf{f}(t_0)\mathbf{f}^T(t_0)]=0$.
  Then, from (\ref{eq:sol}), we obtain that $[y(t_1),\dots,y(t_n)]$ is Gaussian distributed with zero mean and CF
\begin{align}
\nonumber
 &E[y(t_i)y(t_j)]=\tfrac{e^{-\lambda|t_j-t_i|}(1+ \lambda |t_j-t_i|)}{4 \lambda^3}\\
     \nonumber
     &-\tfrac{e^{-\lambda(-2t_0+t_i+t_j)}(1+ \lambda (-2t_0+t_i + t_j) 
     +\lambda^2 (t_0-t_i)(t_0-t_j))}{4 \lambda^3}
\end{align}
for each $t_i,t_j\geq t_0$.
\end{proposition}
It can be observed that for $t_0\rightarrow  -\infty$, the second term goes to zero and
% , thus,
% $$
% E[y(t_i)y(t_j)]\approx \frac{e^{-\lambda|t_j-t_i|}(1+ \lambda |t_j-t_i|)}{4 \lambda^3}
% $$
and we obtain the Mat{\'e}rn CFs for $\nu=3/2$ and $\lambda=\sqrt(3)/\ell$.
Other Mat{\'e}rn CFs for $\nu=3/2,5/2,\dots$ can be obtained via LTI SS models in a similar way.
This connection between LTI SS models and Mat{\'e}rn kernel has been firstly discussed in \cite{sarkka2013spatiotemporal}.
Here, we have reported the CF for $t_0$ finite (non-stationary case) and shown that the CF becomes
stationary for $t_0\rightarrow  -\infty$.

\subsection{Square Exponential kernel}
\label{sec:SE}
Let us now consider the stationary square exponential kernel $E[y(t_i)y(t_j)]=k(\tau)=\exp(-\tfrac{\tau^2}{2\ell^2})$
for $\tau=t_j-t_i$ \citep[Sec. 4.2.1]{rasmussen2006gaussian}.
Its bilateral Laplace transform is  $ S_y(s)= \sqrt{2\pi }\ell  \exp(\tfrac{\ell^2 s^2}{2})$, whose ROC
is $\text{Re}(s)=0$ and, hence, $s=i\omega$.
% Hence, we have that
% $$
%  S_y(i\omega)=\exp(-\tfrac{\ell^2 (\omega)^2}{2})\,\sqrt{2\pi }\ell.
% $$
% and therefore $H(s)H(-s)=\exp(-\tfrac{\ell^2 (\omega)^2}{2})$ and $S_w(s)=\sqrt{2\pi }\ell$.
$ S_y(s)$ is not a rational function, so we cannot directly apply Theorem \ref{th:1}.
However, it is a real-valued positive function (it is positive and non-zero) and it is analytic, so we can approximate $1/S_y(s)$ with  its Taylor expansion
in zero and obtain:
\begin{align}
\label{eq:selaplceb}
 S_y(s)&\approx %\sqrt{2\pi }\ell\frac{1}{1+\frac{1}{2}\ell^2s^2+\dots+\frac{1}{d!2^d}\ell^{2d}s^{2d}}\\%\\
% \label{eq:selaplceb}
 =\sqrt{2\pi }\ell d!2^d\frac{1}{d!2^d+d!2^{d-1}\ell^2s^2+\dots+\ell^{2d}s^{2d}}.
\end{align}
 We can find $H(s)$ by determining the roots of the polynomial at the denominator that have  negative real part.
 This can be done numerically, but the next result is useful.
\begin{theorem}
\label{th:2}
The roots of the denominator in (\ref{eq:selaplceb}) can be obtained by computing the roots for $\ell=1$
and then dividing them for $\ell$. This gives $H(s)$, while $S_w(s)=\sqrt{2\pi }\ell\tfrac{d!2^d}{\ell^{2d}}$.
\end{theorem}
This result is useful because it allows us to compute off-line the solutions of
(\ref{eq:selaplceb}) even when the hyperparameter $\ell$ is unknown.
Let us consider for instance the case $d=2$
$$
S_y(s) \approx 8\sqrt{2\pi }\ell \,\frac{1}{\omega^4\ell^{4} +4\omega^2\ell^{2}+8}.
$$
The roots of the denominator for $\ell=1$ are $s=\pm \sqrt{2\pm i2}$.\footnote{By dividing them for $\ell$ we obtain the roots for $\ell \neq 1$.}
Hence, the roots corresponding to the stable part are $-\sqrt{2\pm i 2}\approx -1.5538\pm i0.6436$ and so
$$
H(s)=\dfrac{1}{(s+a)^2+(b)^2},
$$
with $a=1.5538$ and $b=0.6436$, which can be represented by the following LTI SS model:
  \begin{align}
\nonumber
\mathbf{F}=& \left[\begin{array}{cc}
0  & 1\\
-(a^2+b^2) & -2a\\
 \end{array}\right],~\mathbf{L}= \left[\begin{array}{c}
0 \\
1 \\
 \end{array}\right]\\
\label{eq:sysse}
 \mathbf{C}=& \left[\begin{array}{cc}
1  & 0 \\
 \end{array}\right], 
 \end{align} 
 with $q(t)= 8\sqrt{2\pi }/\ell^3$.
The inverse Laplace transform of $H(s)$ is $h(t)=\tfrac{\exp(-a t) \sin(b t)}{b}$.

 \begin{proposition}
 \label{prop:6}
  Consider (\ref{eq:sysse}) and assume that $E[\mathbf{f}(t_0)\mathbf{f}^T(t_0)]=0$.
  Then, from (\ref{eq:sol}), we obtain that $[y(t_1),\dots,y(t_n)]$ is Gaussian distributed with zero mean and CF
\begin{align}
\nonumber
 E[y(t_i)y(t_j)]&=8\sqrt{2\pi }\\
 %   &\int\limits_{0}^{\min(t_i,t_j)} (t_i-u)e^{-\lambda (t_i-u)}(t_j-u)e^{-\lambda (t_j-u)}du=\\
\nonumber
 &\int\limits_{t_0}^{\min(t_i,t_j)} \frac{\exp(-a (t_i-u)) \sin(b (t_i-u))}{b} \\
 \nonumber
 &\cdot \frac{\exp(-a (t_j-u)) \sin(b (t_j-u))}{b}du
\end{align}
for each $t_i,t_j\geq t_0$. For $t_0\rightarrow -\infty$, it  only depends on $|t_2-t_1|$.
\end{proposition}
This connection between LTI SS models and squared exponential kernels has been derived in \citep{sarkka2013spatiotemporal}
and in \citep{sarkka2014convergence} they have studied the convergence property of the Taylor series.
Here, we have derived the new result in Theorem \ref{th:2}, computed the CF  for a finite $t_0$  (non-stationary case) and shown that the CF becomes
stationary for $t_0\rightarrow  -\infty$ (see appendix for the proof).

\section{LTV systems}
Up to now, we have only worked with LTI SS models. Hereafter, we will show that
moving from LTI to LTV allows us to map two fundamental non-stationary kernels \citep{williams1997computing} to SS models.

\subsection{Non-stationary SE kernel}
Let us consider this CF:
\begin{align}
\label{eq:senonsta}
E[y(t_1)y(t_2)]&=e^{- \frac{1}{2\sigma_m^2}t_i^2} e^{- \frac{1}{2\sigma_s^2}(t_i-t_j)^2} e^{-\frac{1}{2\sigma_m^2}t^2_j}
  \end{align}
  where $\sigma^2_m,\sigma^2_s$ are hyper-parameters.
%   with $1/\sigma_e^2=2/\sigma_g^2+1/\sigma_u^2$, $\sigma_s^2=2\sigma_g^2+\sigma_4^2/\sigma_u^2$ and  $\sigma_m^2=2\sigma_u^2+\sigma_g^2$.
  This CF is clearly non-stationary. However, its central term is the square exponential CF and, therefore, it can be approximated by a LTV SS
  that is a simple variant of the LTI SS that defines the square exponential CF.
  For instance, for $\sigma_s=\sigma_m=1$ this LTV SS model is:
  \begin{align}
\nonumber
\mathbf{F}=& \left[\begin{array}{cc}
0  & 1\\
-(a^2+b^2) & -2a\\
 \end{array}\right],~\mathbf{L}= \left[\begin{array}{c}
0 \\
1 \\
 \end{array}\right]\\
\label{eq:ltvex}
 \mathbf{C}=& \left[\begin{array}{cc}
e^{-t^2/2} & 0 \\
 \end{array}\right], 
 \end{align} 
  The impulse response is  $h(t,u)=e^{-t^2/2} \tfrac{\exp(-a (t-u)) \sin(b (t-u))}{b}$ for $u<t$ 
and the CF
\begin{align}
\nonumber
 E[y(t_i)y(t_j)]&=8\sqrt{2\pi }e^{-t_1^2/2}e^{-t_2^2/2}\\
%   &\int\limits_{0}^{\min(t_i,t_j)} (t_i-u)e^{-\lambda (t_i-u)}(t_j-u)e^{-\lambda (t_j-u)}du=\\
\nonumber
 & \int\limits_{t_0}^{\min(t_i,t_j)} \frac{\exp(-a (t_i-u)) \sin(b (t_i-u))}{b} \\
 \nonumber
 &\cdot \frac{\exp(-a (t_j-u)) \sin(b (t_j-u))}{b}du
\end{align}
This is the $d=2$ approximation.
For $t_0 \rightarrow -\infty$, the integral term depends only on $|t_2-t_1|$.
This CF is used in neural networks research and the connection with GPs 
has been discussed by \cite{williams1997computing}.

\subsection{LTV system defining the neural network kernel}
Let us consider the following LTV SS model
  \begin{align}
\label{eq:ltvex}
\mathbf{F}=& \left[\begin{array}{cc}
0  & e^{-\mathbf{t}^T\boldsymbol{\Sigma}^{1/2}\mathbf{r}}\\
0 & 0\\
 \end{array}\right], \mathbf{C}= \left[\begin{array}{cc}
1 & 0 \\
 \end{array}\right],
 \end{align} 
 with $q(t)=0 \forall t>0$, $\mathbf{t}=[1,t]^T$, $\mathbf{r}=[r_0,r_1]^T$
 and $\boldsymbol{\Sigma}=diag(\sigma_0^2,\sigma_1^2)$ is a definite positive definite matrix.
Its  transfer function is
 $$
 \phi(t_i,t_0)= \left[\begin{array}{cc}
1  & \tfrac{\sqrt{\pi } (\text{erf}(\mathbf{t_i}^T\boldsymbol{\Sigma}^{1/2}\mathbf{r})-\text{erf}(\mathbf{t}_0^T\boldsymbol{\Sigma}^{1/2}\mathbf{r}))}{2 {r_1 \sigma_1}}\\
0 & 1\\
 \end{array}\right].
 $$
 Hence,  we have that
  \begin{equation}
\label{eq:serfmodel}
\begin{array}{l}
 y(t_i)=\mathbf{C}\boldsymbol{\psi}(t_i,t_0)\mathbf{f}(t_0)\\
 =f_1(t_0)+\tfrac{\sqrt{\pi } (\text{erf}(\mathbf{t_i}^T\boldsymbol{\Sigma}^{1/2}\mathbf{r})-\text{erf}(\mathbf{t}_0^T\boldsymbol{\Sigma}^{1/2}\mathbf{r}))}{2 {r_1 \sigma_1}} f_2(t_0)\end{array}
\end{equation}
 and
  \begin{equation}
\label{eq:solcovi0}
\begin{array}{l}
E[y(t_i)y(t_j)]=\vspace{1mm}\\
=\mathbf{C}(t_i)\boldsymbol{\psi}(t_i,t_0)E[\mathbf{f}(t_0)\mathbf{f}^T(t_0)](\mathbf{C}(t_j)\boldsymbol{\psi}(t_j,t_0))^T \vspace{1mm}\\
=\text{erf}(\mathbf{t}_i^T\boldsymbol{\Sigma}^{1/2}\mathbf{r})\text{erf}(\mathbf{t}_j^T\boldsymbol{\Sigma}^{1/2}\mathbf{r}),
\end{array}
\end{equation}
where  we have assumed that $\mathbf{f}(t_0)$ is Gaussian distributed with zero mean and covariance matrix
$E[\mathbf{f}(t_0)\mathbf{f}^T(t_0)]=[\text{erf}(\mathbf{t}_0^T\boldsymbol{\Sigma}^{1/2}\mathbf{r}), \tfrac{2 r_1 \sigma_1}{\sqrt{2\pi}}]^T[\text{erf}(\mathbf{t}_0^T\boldsymbol{\Sigma}^{1/2}\mathbf{r}), \tfrac{2 r_1 \sigma_1}{\sqrt{2\pi}}]$.
Therefore, from (\ref{eq:serfmodel}), it is evident that (\ref{eq:ltvex}) models $\text{erf}$ like functions.
% Figure \ref{fig:erf} shows some trajectories generated by this model and the contour plot of the CF.
% \begin{figure}
% \centering
%  \includegraphics[width=5cm]{Plots_Paper_Erf.pdf}
% %   \includegraphics[width=4.5cm]{Contour_Erf.pdf}
% \caption{Random plots generated according to (\ref{eq:serfmodel}).}
%    \label{fig:erf}
% \end{figure}
We can add expressivity to the model as follows 
  \begin{align}
\nonumber
\mathbf{F}=& \left[\begin{array}{cccc}
0  & e^{-\mathbf{t}^T\boldsymbol{\Sigma}^{1/2}\mathbf{r}_1} & 0 & 0\\
0 & 0& 0 & 0\\
0& 0 & 0  & e^{-\mathbf{t}^T\boldsymbol{\Sigma}^{1/2}\mathbf{r}_2} \\
0 & 0& 0 & 0\\
 \end{array}\right],\\ \nonumber \mathbf{C}=& \left[\begin{array}{cccc}
1 & 0 & 1 & 0\\
 \end{array}\right]/\sqrt{2}.
 \end{align} 
Its  transfer function $\phi(t_i,t_0)$ is a diagonal block matrix with elements
 $$
\left[\begin{array}{cc}
1  & \tfrac{\sqrt{\pi } (\text{erf}(\mathbf{t_i}^T\boldsymbol{\Sigma}^{1/2}\mathbf{r}_k)-\text{erf}(\mathbf{t}_0^T\boldsymbol{\Sigma}^{1/2}\mathbf{r}_k))}{2 {r_{1k} \sigma_1}}\\
0 & 1\\
 \end{array}\right],
 $$
 for $k=1,2$.
 Hence, we have that
   \begin{equation}
\label{eq:serfmodel1}
\begin{array}{l}
 y(t_i)=\\
 \frac{1}{\sqrt{2}}\sum\limits_{k=1}^2 f_{1k}(t_0)+\tfrac{\sqrt{\pi } (\text{erf}(\mathbf{t_i}^T\boldsymbol{\Sigma}^{1/2}\mathbf{r}_k)-\text{erf}(\mathbf{t}_0^T\boldsymbol{\Sigma}^{1/2}\mathbf{r}_k))}{2 {r_{1k} \sigma_1}} f_{2k}(t_0),\end{array}
\end{equation}
 where $\mathbf{f}(t_0)=[f_{11}(t_0),f_{21}(t_0),f_{12}(t_0),f_{22}(t_0)]^T$ and so
 \begin{equation}
\label{eq:solcovi01}
\begin{array}{l}
E[y(t_i)y(t_j)]=\vspace{1mm}\\
=\mathbf{C}(t_i)\boldsymbol{\psi}(t_i,t_0)E[\mathbf{f}(t_0)\mathbf{f}^T(t_0)](\mathbf{C}(t_j)\boldsymbol{\psi}(t_j,t_0))^T \vspace{1mm}\\
=\frac{1}{2}\sum\limits_{k=1}^2 \text{erf}(\mathbf{t}_i^T\boldsymbol{\Sigma}^{1/2}\mathbf{r}_k)\text{erf}(\mathbf{t}_j^T\boldsymbol{\Sigma}^{1/2}\mathbf{r}_k)
\end{array}
\end{equation}
where we have assumed that $E[\mathbf{f}(t_0)\mathbf{f}^T(t_0)]$ is a block diagonal matrix with elements 
$[\text{erf}(\mathbf{t}_0^T\boldsymbol{\Sigma}^{1/2}\mathbf{r}_k), \tfrac{2 r_{1k} \sigma_1}{\sqrt{2\pi}}]^T[\text{erf}(\mathbf{t}_0^T\boldsymbol{\Sigma}^{1/2}\mathbf{r}_k), \tfrac{2 r_{1k} \sigma_1}{\sqrt{2\pi}}]$.
If we increase the number of elements $k$ and keep this block-diagonal structure, we have that
  \begin{equation}
\label{eq:solcovi02}
\begin{array}{l}
E[y(t_i)y(t_j)]=\frac{1}{J}\sum\limits_{k=1}^J \text{erf}(\mathbf{t}_i^T\boldsymbol{\Sigma}^{1/2}\mathbf{r}_k)\text{erf}(\mathbf{t}_j^T\boldsymbol{\Sigma}^{1/2}\mathbf{r}_k).
\end{array}
\end{equation}

\begin{figure}
\centering
 \includegraphics[width=5cm]{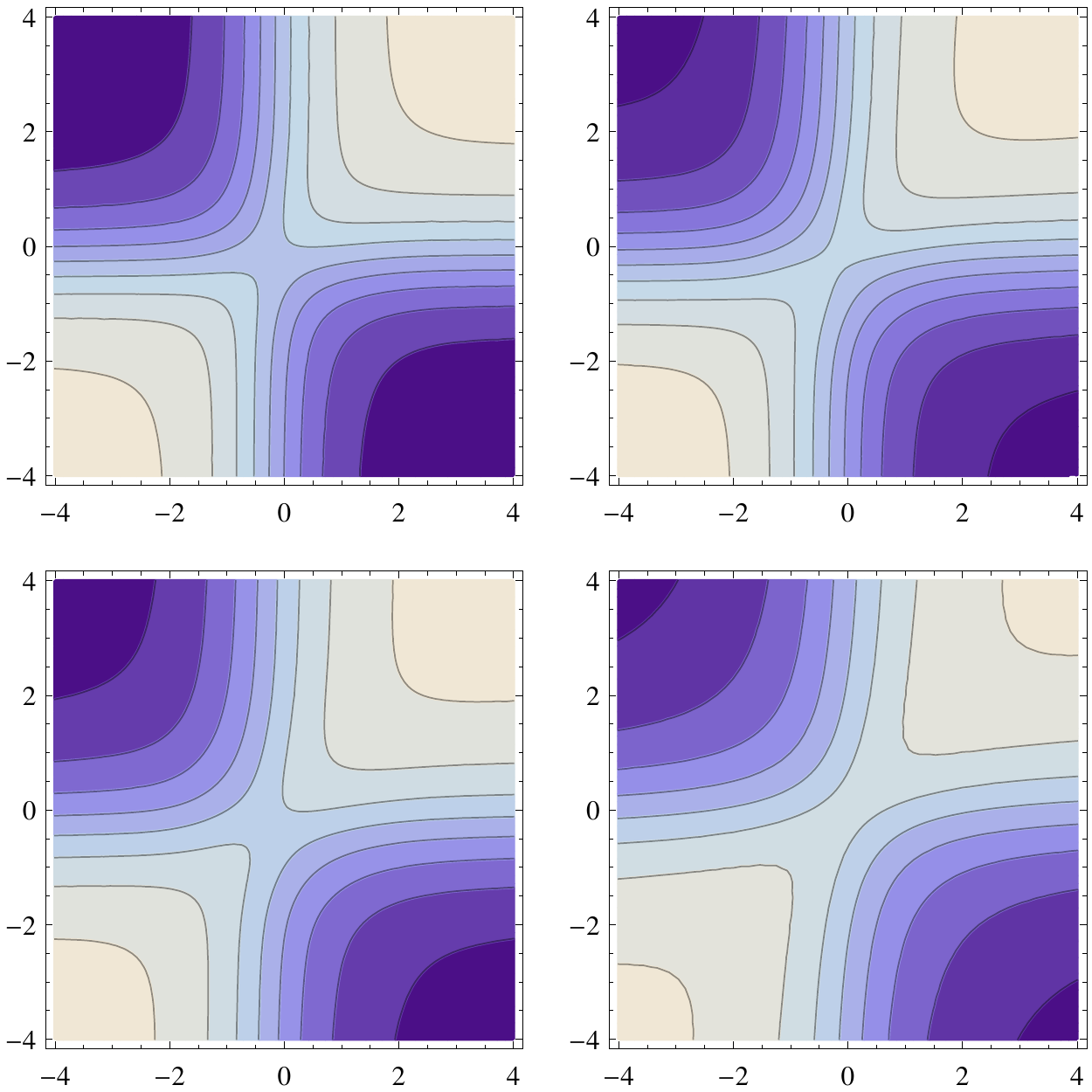}
 \caption{Contour plot of the NN CF for $J=15,30,50,\infty$ (from left to right).}
 \label{fig:erf4}
\end{figure}

\begin{figure}
\centering
 \includegraphics[width=5cm]{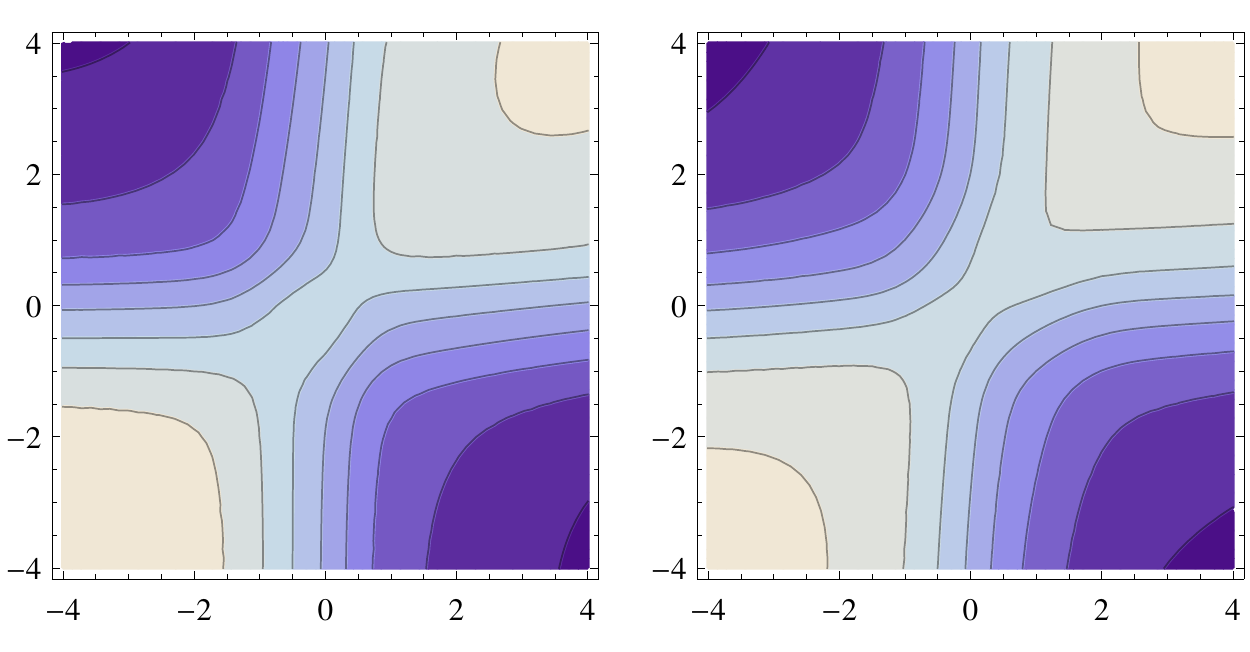}
 \caption{Contour plot of the NN CF using deterministic sampling for $J=10,20$ (from left to right).}
 \label{fig:erf2}
\end{figure}

\begin{theorem}
\label{th:3}
 Assume that the vectors $\mathbf{r}_k$ are sampled from a standard Gaussian distribution, then 
   \begin{equation}
\label{eq:solcovi02gen}
   \begin{array}{l}
E[y(t_i)y(t_j)]=\frac{1}{J} \sum\limits_{k=1}^J \text{erf}(\mathbf{t}_i^T\boldsymbol{\Sigma}^{1/2}\mathbf{r}_k)\text{erf}(\mathbf{t}_j^T\boldsymbol{\Sigma}^{1/2}\mathbf{r}_k)\vspace{1mm}\\
\xrightarrow{J \rightarrow \infty} \int \text{erf}(\mathbf{t}_i^T\boldsymbol{\Sigma}^{1/2}\mathbf{r})\text{erf}(\mathbf{t}_j^T\boldsymbol{\Sigma}^{1/2}\mathbf{r})N(\mathbf{r};0,\mathbf{I})d\mathbf{r}\vspace{1mm}\\
\propto %\int \text{erf}(\mathbf{t}_i^T\mathbf{r})\text{erf}(\mathbf{t}_j^T\mathbf{r})N(\mathbf{r};0,\boldsymbol{\Sigma})d\mathbf{r}\\
\frac{2}{\pi} \sin^{-1} \left(\frac{2 \mathbf{t}_i^T \boldsymbol{\Sigma} \mathbf{t}_j}{\sqrt{(1+2 \mathbf{t}_i^T \boldsymbol{\Sigma} \mathbf{t}_i )(1+2 \mathbf{t}_j^T \boldsymbol{\Sigma} \mathbf{t}_j )}}\right)
\end{array}
\end{equation}
which is the neural network CF \citep{williams1997computing}.
\end{theorem}
Figure \ref{fig:erf4} shows the contourplot of the CF for different $J$.
It is evident that at the increase of $J$ converges to the NN CF.
To increase the convergence speed, we can choose the vectors $\mathbf{r}_k$ in a deterministic way (smart sampling).
We follow the approach proposed by \cite{huber2008gaussian}. Figure \ref{fig:erf2} shows the contourplot of the CF for
the deterministic sampling. it is evident that for $J=20$ the approximation is already very good (compare it with the last plot in Figure \ref{fig:erf4}
relative to $J=\infty$).

\section{Scaling and measurement noise}
In the previous derivations, we have assumed that there is not measurement noise.
We can add a measurement noise to the measurement equation as follows
\begin{equation}
 y(t_i)=\mathbf{C}(t_i)\mathbf{x}(t_i)+v(t_i),
\end{equation}
where $v(t_i)=N(0,\sigma_v^2)$ is assumed to be independent to the process noise and
initial condition. Under these assumptions, for each CF $=(\dots)$ computed in the above sections,  we have now that
$$
E[y(t_i)y(t_j)]=(\dots) + \sigma_v^2.
$$
Moreover, sometimes it is desirable to have a positive scaling parameter that multiplies the CF.
This can be included by rescaling either the stochastic forcing term  or the initial condition (for SS without forcing term).
Finally, note that different kernels can be combined in an additive way, by simply stacking SS models in block-diagonal matrices
as shown for the periodic or NN SS model.

\section{Estimates}
Once we have defined the SS model for a given CF, we aim to estimate
the state of the model given the measurements.
Given the hyperparameters of the CF, this can be carried out in three steps:\\
(a) Discretize the continuous-time SDE to obtain a 
 discrete-time SDE. This step basically consists on applying (\ref{eq:sol}).\\ 
 (b) Compute  the probability density function (PDF)  $p(\mathbf{x}(t_k)|y(t_1),\dots,y(t_k))$, that is Gaussian.
 The mean and covariance  matrix of this Gaussian PDF can be computed efficiently by using Kalman filtering. \\ 
 (c) Compute  the Gaussian posterior PDF  $p(\mathbf{x}(t_k)|y(t_1),\dots,y(t_n))$.
 The mean and covariance  matrix of this PDF can be computed very efficiently by smoothing the estimates
 obtained by the Kalman filter.\\
 The last step return the estimates of the state given all observations.
The last two steps have both complexity $\mathcal{O}(n)$. To estimate the hyperparameters
of the CF, we adopt a Bayesian approach. We place a prior on the hyperparameters
and then we estimate them using a Monte Carlo approach. In practice, we employ a Rao-Blackwellised particle filtering
\cite[Ch. 24]{smith2013sequential}.
We have reported the steps of the whole inference scheme in appendix.

\section{Example sawtooth wave and Gaussian}
\begin{figure}[h]
\centering
 \includegraphics[width=4cm]{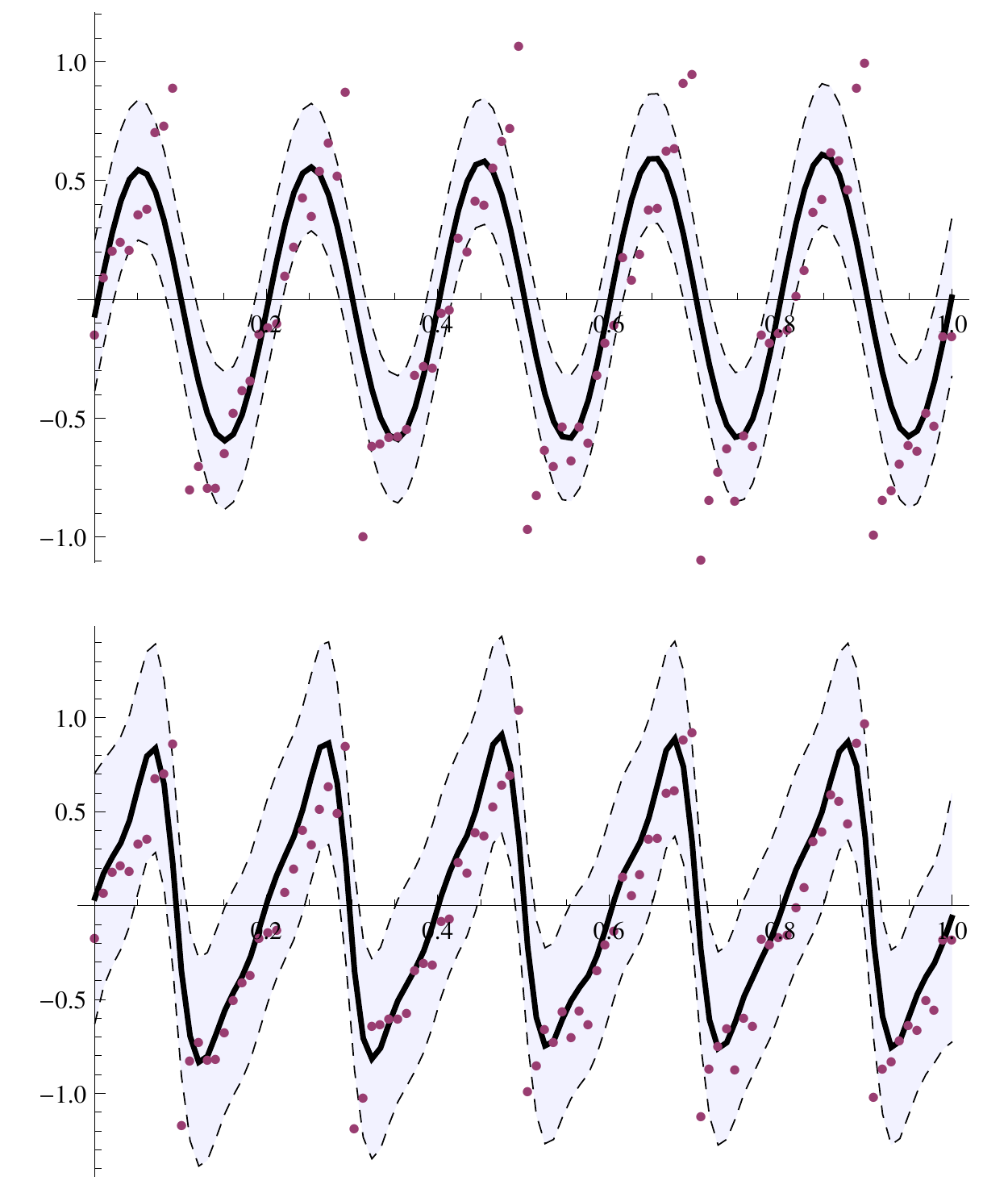}
  \includegraphics[width=4cm]{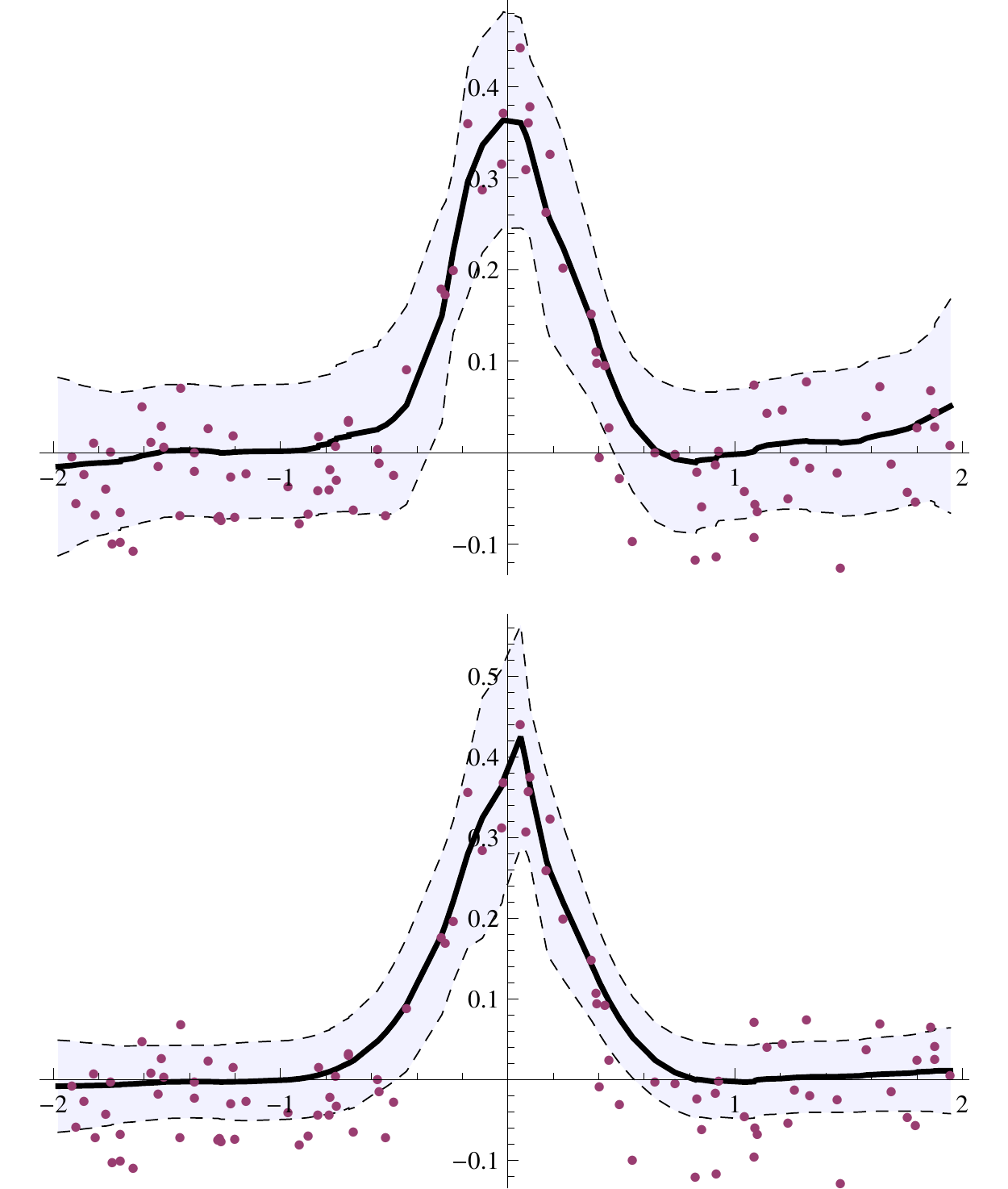}
 \caption{Sawtooth inference with SS periodic model of order $J=2,6$. Gaussian inference with SS NN model for $J=10,20$.}
 \label{fig:periodic_order}
\end{figure}

Consider the following two  functions
$$
y_a(t)=S(10\pi(t+0.1))+v_a(t), ~~y_b(t)=N(4t,0,1)+v_b(t)
$$
with $S(t)=t-\lfloor t \rfloor$, $v_a(t) \sim N(0,0.1)$ and $v_b(t) \sim N(0,0.05)$. The first is a sawtooth wave with period $0.2$. The second is a Gaussian density.
 We have generated $100$ observation from the above regression  model considering $t \in [0:0.01:1]$ in the first case 
 and randomly generated $t\sim 4 Unif[0,1] -2$ in the second case.
Our goal is to employ  the periodic SS model and, respectively, NN SS model to
estimate the two functions.
The result  is shown in Figure  \ref{fig:periodic_order}.
In particular, the two paired plots report the posterior mean and relative credible region
for two different orders of the periodic and NN SS model.
Consider the sawtooth case. The first plot shows the estimate based on the periodic SS model of order one (only one cosine
component). This SS model can estimate exactly only periodic cosine-type function, but the unknown function is a periodic sawtooth wave.
To improve the estimate, we can increase the number of cosine terms. For $J=6$, it is evident that we already obtain an 
accurate estimate of the sawtooth wave. 
A similar comment holds also for the  NN SS model. At the increasing of the approximation order, the SS posterior estimate 
gets closer to the true function (it better models the tails of the Gaussian PDF that has generated the data).

% \begin{figure}[h]
% \centering
% 
%  \caption{Gaussian inference with SS NN model for $J=10,20$.}
%  \label{fig:nn_order}
% \end{figure}

\begin{figure}[h]
\centering
 \includegraphics[width=5cm]{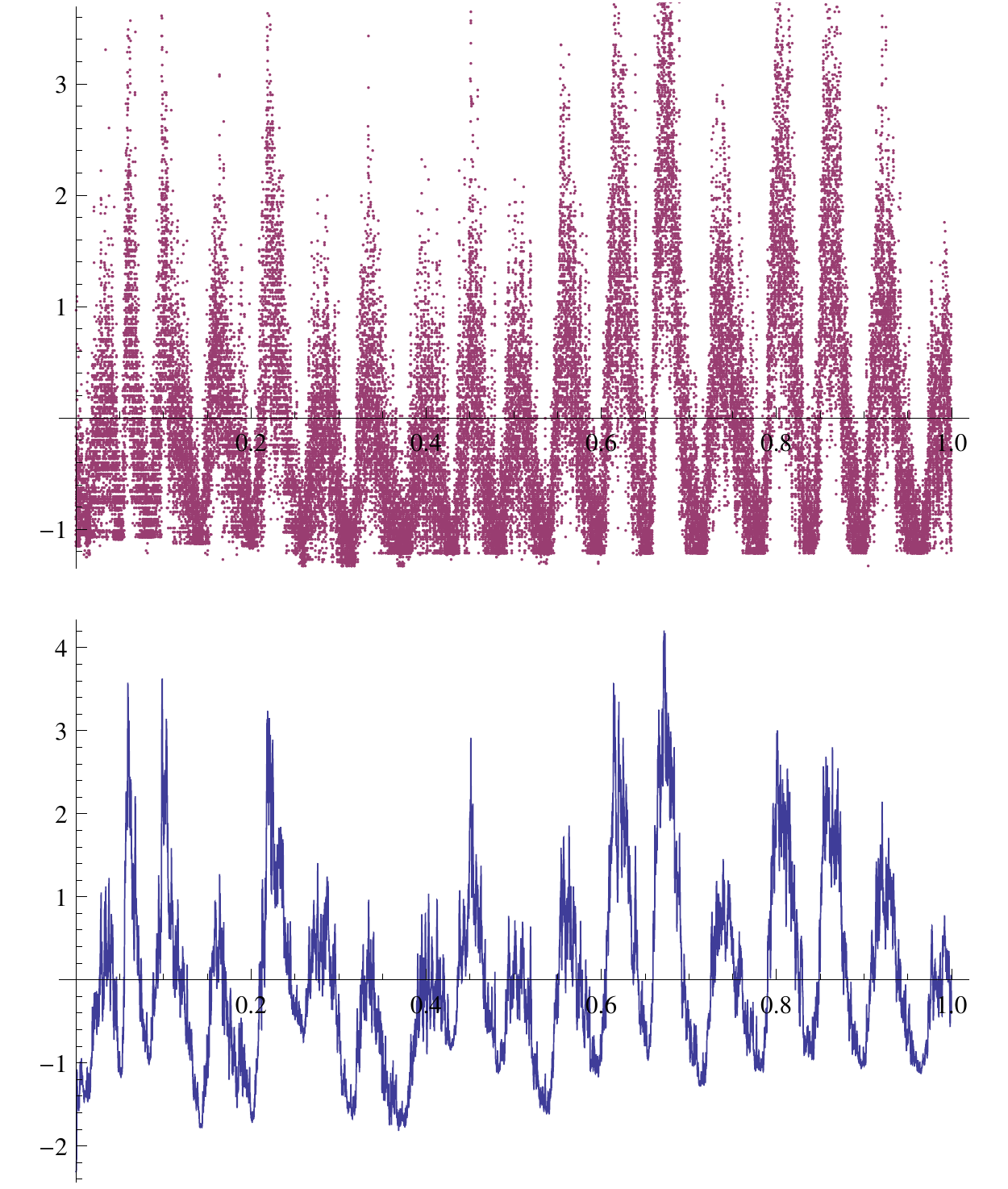}
 \caption{Sunspots number}
 \label{fig:sunspots1}
\end{figure}

\subsection{Big Data: sunspots number}
Finally, to show the computational efficiently  of the proposed  approach, we have applied
the SS model for inferences in a long time-series.
In particular, we have considered the sunspot time series -- a daily time series that reports the number of sunspots from 1820 to 2015.
After treated the missing data, the number of observations is $58'306$.
For inferences we have applied the periodic SS models with the goal of estimating
the period of sunspot activity cycle, that is about eleven years.
Figure \ref{fig:sunspots1} reports the observations (the time has been normalized in $[0,1]$ and 
the number of sunspots have been standardized) and the posterior mean computed using the periodic SS model
with $J=8$ components.
The estimated period  for the sunspot cycle was $10.36$ years. The time to run the SS model on this time series
was $1.5$ hours on standard laptop with a model implemented in Matlab.

\section{Conclusions}
In this paper we have shown that, by exploiting the transient behaviour of SS models, it is possible to map non-stationary Gaussian Processes (GP) kernels
to state space models (SS). In particular, we have shown how to map to SS models the neural network Kernels.
This is important because SS models allows to reduce the computational complexity of inferences with GPs from cubic to linear in the number of observations.
As future work, we plan to continue to study the relationship between the SS model representation and GPs as well as we plan to 
extend this work to multivariate regression problems.

% \subsubsection*{Acknowledgements}
% 
% Use unnumbered third level headings for the acknowledgements.  All
% acknowledgements go at the end of the paper.  Be sure to omit any
% identifying information in the initial double-blind submission!

\appendix

\bibliography{biblio1}
\bibliographystyle{plainnat}

\newpage 

  \appendix
  \section{Proofs}
  
  \subsection{Proposition \ref{prop:1}}
By definition of covariance, we have that
\begin{equation}
\begin{array}{l}
E[y(t_i)y(t_j)]=\vspace{1mm}\\
\mathbf{C}(t_i)\boldsymbol{\psi}(t_i,t_0)E[\mathbf{f}(t_0)\mathbf{f}^T(t_0)] (\mathbf{C}(t_j)\boldsymbol{\psi}(t_j,t_0))^T+\vspace{1mm}\\
\int\limits_{t_0}^{t_i} \int\limits_{t_0}^{t_j}  \mathbf{C}(t_i) \boldsymbol{\psi}(t_i,u)\mathbf{L}(u)\,E[dw(u)dw(v)]\\
\cdot \mathbf{L}^T(v)\boldsymbol{\psi}^T(t_j,v)\mathbf{C}^T(t_j) \vspace{1mm}\\
=\mathbf{C}(t_i)\boldsymbol{\psi}(t_i,t_0)E[\mathbf{f}(t_0)\mathbf{f}^T(t_0)] (\mathbf{C}(t_j)\boldsymbol{\psi}(t_j,t_0))^T+\vspace{1mm}\\
\int\limits_{t_0}^{\min(t_i,t_j)}  \mathbf{C}(t_i) \boldsymbol{\psi}(t_i,u)\mathbf{L}(u)\mathbf{L}^T(u)\boldsymbol{\psi}^T(t_j,u)\mathbf{C}^T(t_j)q(u)du\vspace{1mm}\\
=\mathbf{C}(t_i)\boldsymbol{\psi}(t_i,t_0)E[\mathbf{f}(t_0)\mathbf{f}^T(t_0)] (\mathbf{C}(t_j)\boldsymbol{\psi}(t_j,t_0))^T+\vspace{1mm}\\
\int\limits_{t_0}^{\min(t_i,t_j)} h(t_i,u)h(t_j,u)q(u)du\vspace{1mm}\\
\end{array}
\end{equation} 
where we have exploited that $E[dw(u)dw(v)]=q(u)\delta(u-v)dudv$ and defined $h(t,u)=\mathbf{C}(t) \boldsymbol{\psi}(t,u)\mathbf{L}(u)$.

  \subsection{Proposition \ref{prop:2}}
Since $y(t_i)=\mathbf{C}\boldsymbol{\psi}(t_i,t_0)\mathbf{f}(t_0)=f_1(t_0)+(t-t_0) f_2(t_0)$.
and $\mathbf{f}(t_0)$ is Gaussian distributed with zero mean and covariance $E[\mathbf{f}(t_0)\mathbf{f}^T(t_0)]=diag([\sigma_1^2,\sigma_2^2])$, from (\ref{eq:sol}),
we have that for $t_0=0$:
\begin{equation}
\begin{array}{l}
E[y(t_i)y(t_j)]
=E[ (f_1(0)+t_i f_2(0))(f_1(0)+t_j f_2(0))] \\
= \sigma_0^2 +t_i t_j\sigma_1^2
\end{array}
\end{equation}
 for each $t_i,t_j\geq 0$. 
 
     \subsection{Proposition \ref{prop:3}}
     The proof is similar to that of Proposition \ref{prop:2} but 
   considering that
     \begin{align}
  \nonumber
  y(t_i)&=\mathbf{C}\boldsymbol{\psi}(t_i,t_0)\mathbf{f}(t_0)\\
\nonumber
&=\cos(\omega_k (t-t_0))f_1(t_0)+ \sin(\omega_k (t-t_0))f_2(t_0)\\
    \end{align} 
  
    \subsection{Proposition \ref{prop:4}}
    In this case, we have that $h(t,u)=h(t-u)=(t-u)$ and so from (\ref{eq:sol}),
    \begin{align}
\nonumber
 E[y(t_i)y(t_j)]&=\int\limits_{t_0}^{\min(t_i,t_j)} h(t_i,u)h(t_j,u)q(u)du\\
 \nonumber
 &\int\limits_{0}^{\min(t_i,t_j)} (t_i-u)(t_j-u)du,\\
\nonumber
 &=|t_i-t_j|\frac{\min(t_i,t_j)^2}{2}+\frac{\min(t_i,t_j)^3}{3},
\end{align}
where we have exploited the fact that $q(u)=1$.

    \subsection{Proposition \ref{prop:5}}
\begin{align}
\nonumber
 &E[y(t_i)y(t_j)]=\\
%   &\int\limits_{0}^{\min(t_i,t_j)} (t_i-u)e^{-\lambda (t_i-u)}(t_j-u)e^{-\lambda (t_j-u)}du=\\
\nonumber
 &\int\limits_{t_0}^{\min(t_i,t_j)} (t_i-u)e^{-\lambda (t_i-u)}(t_j-u)e^{-\lambda (t_j-u)}du\\
\nonumber
 &\tfrac{e^{-\lambda|t_j-t_i|}(1+ \lambda |t_j-t_i|)}{4 \lambda^3}\\
     \nonumber
     &-\tfrac{e^{-\lambda(-2t_0+t_i+t_j)}(1+ \lambda (-2t_0+t_i + t_j) 
     +\lambda^2 (t_0-t_i)(t_0-t_j))}{4 \lambda^3}
\end{align}
for each $t_i,t_j\geq t_0$.
It can be observed that for $t_0\rightarrow  -\infty$, the second term goes to zero.

    \subsection{Proposition \ref{prop:6}}
    The result of the integral is
\begin{align}
\nonumber
&\Big(e^{-a (-2 \min (t_i,t_j)+t_i+t_j)}
   \Big(a (b \sin (b (-2 \min
   (t_i,t_j)+t_i+t_j))\\
   \nonumber
   &-a \cos (b (-2 \min
   (t_i,t_j)+t_i+t_j)))+\left(a^2+b^2\right) \cos (b (t_i-t_j))\Big)\\
   \nonumber
   &-e^{-a (-2   t_0+t_i+t_j)} \Big(\left(a^2+b^2\right) \cos
   (b (t_i-t_j))\\
   \nonumber
   &+a (b \sin (b (-2
   t_0+t_i+t_j))-a \cos (b (-2
   t_0+t_i+t_j)))\Big)\Big)\\
   \nonumber
   &\left(4 a b^2    \left(a^2+b^2\right)\right)^{-1}
\end{align}
The second term that multiplies $e^{-a (-2    t_0+t_i+t_j)}  $ vanishes for $t_0\rightarrow -\infty$ and so we have 
\begin{align}
\nonumber
&e^{-a |t_j-t_i|}
   \Big(a (b \sin (b |t_j-t_i|)\\
   \nonumber
   &-a \cos (b |t_j-t_i|)+\left(a^2+b^2\right) \cos (b |t_j-t_i|)\Big)\\
       \nonumber
    &\left(4 a b^2\left(a^2+b^2\right)\right)^{-1}
\end{align}
where we have exploited the fact that $-2 \min     (t_i,t_j)+t_i+t_j=|t_j-t_i|$
and cosine is an even function.

\subsection{Theorem \ref{th:2}}
This result is obvious by noticing that in
\begin{align}
\nonumber
\sqrt{2\pi }\ell d!2^d\frac{1}{d!2^d+d!2^{d-1}\ell^2s^2+\dots+\ell^{2d}s^{2d}}
\end{align}
the variable $s^o$ always appears multiplied by $\ell^o$  with the same power $o$.
Then by redefining $x=s\ell$ we can prove the theorem.
  
\subsection{Theorem \ref{th:3}}  
Consider 
   \begin{equation}
\nonumber
   \begin{array}{l}
E[y(t_i)y(t_j)]=\frac{1}{J} \sum\limits_{k=1}^J \text{erf}(\mathbf{t}_i^T\boldsymbol{\Sigma}^{1/2}\mathbf{r}_k)\text{erf}(\mathbf{t}_j^T\boldsymbol{\Sigma}^{1/2}\mathbf{r}_k)\vspace{1mm}\\
\xrightarrow{J \rightarrow \infty} \int \text{erf}(\mathbf{t}_i^T\boldsymbol{\Sigma}^{1/2}\mathbf{r})\text{erf}(\mathbf{t}_j^T\boldsymbol{\Sigma}^{1/2}\mathbf{r})N(\mathbf{r};0,\mathbf{I})d\mathbf{r}\vspace{1mm}\\
\propto %\int \text{erf}(\mathbf{t}_i^T\mathbf{r})\text{erf}(\mathbf{t}_j^T\mathbf{r})N(\mathbf{r};0,\boldsymbol{\Sigma})d\mathbf{r}\\
\frac{2}{\pi} \sin^{-1} \left(\frac{2 \mathbf{t}_i^T \boldsymbol{\Sigma} \mathbf{t}_j}{\sqrt{(1+2 \mathbf{t}_i^T \boldsymbol{\Sigma} \mathbf{t}_i )(1+2 \mathbf{t}_j^T \boldsymbol{\Sigma} \mathbf{t}_j )}}\right)
\end{array}
\end{equation}
the first equality follows from the Strong Law of Large Numbers. The second equality can be proven by a change of variables $\mathbf{z}=\boldsymbol{\Sigma}^{1/2}\mathbf{r}$
and exploiting a result proven by  \cite{williams1997computing}.

\def\trans {{^T}}
\section{Kalman filtering and smoothing}
Consider the discretized linear system
\begin{align}
x_{t+1} &= Ax_{t} + L w_{t}\notag\\
y_{t} &= Cx_{t} + v_{t}
\label{eqn:system}
\end{align}
where $w_{t} \sim N\left(0,\Sigma_{w}\right)$, $v_{t} \sim
N\left(0,\Sigma_{v}\right)$, and $x_{0} \sim
N\left(x_{0|-1},P_{0|-1}\right)$. Note that $x \sim
N\left(\mu,\Sigma\right)$ means

\begin{equation}
P\left(x\right) = \frac{1}{\left(2\pi\right){|\Sigma|}^{1/2}} e^{-\frac{1}{2}\left(x-\mu\right)\trans\Sigma^{-1}\left(x-\mu\right)}.\notag
\end{equation}
We also have: $E(x) = \mu$ and $E\left(x-\mu\right)\left(x-\mu\right)\trans = \Sigma$.

We will use the following notation:
\begin{align}
\hat{x}_{t|t} &= E\left[x_{t}|y_{0\colon t}\right]\notag\\
P_{t|t} &= E\left[\left(x_{t} - \hat{x}_{t|t}\right)\left(x_{t} - \hat{x}_{t|t}\right)\trans|y_{0\colon t}\right]\notag\\
\hat{x}_{t+1|t} &= E\left[x_{t+1}|y_{0\colon t}\right]\notag\\
P_{t+1|t} &= E\left[\left(x_{t+1} - \hat{x}_{t+1|t}\right)\left(x_{t+1} - \hat{x}_{t+1|t}\right)\trans|y_{0\colon t}\right]\notag
\end{align}

Note that because $x_{t|\cdot}$ is a Gaussian random variable, it is
sufficient to only keep track of the mean and covariance.  We can do
so by the following computations at each time $t$:
\begin{align}
\hat{x}_{t+1|t} &= A\hat{x}_{t|t} + Bu_{t}\notag\\
P_{t+1|t} &= AP_{t|t}A\trans +\Sigma_{w}\notag\\
\hat{x}_{t+1|t+1} &= \hat{x}_{t+1|t} + K_{t+1}\left(y_{t+1}-C\hat{x}_{t+1|t}\right)\notag\\
K_{t+1} &= P_{t+1|t}C\trans\left(CP_{t+1|t}C\trans + \Sigma_{v}\right)^{-1}\notag\\
P_{t+1|t+1} &= P_{t+1|t} - P_{t+1|t}C\trans\left(CP_{t+1|t}C\trans + \Sigma_{v}\right)^{-1}CP_{t+1|t}
\end{align}
These are the equations of the Kalman filter and return $p(x(t_k)|y(t_1),\dots,y(t_k))$.

\subsection{Kalman Smoother}
The filtered estimate of $x(t_k)$ only takes into account the
``past'' information relative to $x(t_k)$.  By incorporating the
``future'' observations relative to $x(t_k)$, we can obtain a more
refined state estimate.  

Estimators that take into account both past and future are often
called ``smoothers.''  The Kalman smoother estimates
$p(x(t_k)|y(t_1),\dots,y(t_n))$.  

The Kalman smoother equations here:

\begin{align}
\hat{x}_{t|T} &= \hat{x}_{t|t} + L_{t}\left(x_{t+1|T}-\hat{x}_{t+1|T}\right) \label{eqn:smoother-x}\\
P_{t|T} &= P_{t|t} + L_{t}\left(P_{t+1|T}-P_{t+1|t}\right)L_{t}\trans \label{eqn:smoother-cov} \\
L_{t} &= P_{t|t}A\trans P_{t+1|t}^{-1}
\end{align}

Before running the smoother, we must first run the filter.
The smoother then proceeds backward in time.

Note that $P_{t+1|T}-P_{t+1|t} < 0$ as the uncertainty over $x_{t+1}$
is smaller when conditioned on all observations, than when only
conditioned on past observations.

  \subsection{Particle filter}
  The Kalman filter and Smoother assume that the matrices $A,C,L$, the variances of the process and measurement noises
  are completely known.
  In the GPs, they may include the hyperparameters that we aim to estimate from data.
  There are many ways to estimate these parameters. 
We use a pure Bayesian approach. We place a prior on the hyperparameters
and then we estimate them using a Monte Carlo approach. In practice, we employ a Rao-Blackwellised particle filtering
\cite[Ch. 24]{smith2013sequential}, by exploiting the fact that the system is linear given the hyperparameters and so we can run the Kalamn filter
and Smoother and then update the hyperparameters.
It is exactly like computing inferences in  a hierarchical model.
In all the experiments we have used a uniform prior for the hyperparameters in $[0,10]$.

\end{document}